# Does it care what you asked?
# Understanding Importance of Verbs in Deep Learning QA System

Barbara Rychalska[*1], Dominika Basaj[*2], Przemysław Biecek[3], and Anna Wróblewska[4]

d.basaj, b.rychalska, p.biecek, a.wroblewska@mini.pw.edu.pl
[1,2,3,4]Warsaw University of Technology, Warsaw, Poland
[2]Tooploox, Wrocław, Poland

## Abstract

In this paper we present the results of an investigation of the importance of verbs in a deep learning QA system trained on SQuAD dataset. We show that main verbs in questions carry little influence on the decisions made by the system - in over 90% of researched cases swapping verbs for their antonyms did not change system decision. We track this phenomenon down to the insides of the net, analyzing the mechanism of self-attention and values contained in hidden layers of RNN. Finally, we recognize the characteristics of the SQuAD dataset as the source of the problem. Our work refers to the recently popular topic of adversarial examples in NLP, combined with investigating deep net structure.

## 1 Introduction

Recent advances in interpretability for NLP focus on the problem of adversarial examples (Ribeiro et al., 2018) (Jia and Liang, 2017) which lead systems to mistakenly change output. In case of QA systems, either questions or contexts are modified, and it is shown that seemingly small changes in semantics flip system decisions.

In this paper we take a different approach: we create heavy differences in meaning by generating questions with their meaning negated, and observe system outputs. Our initial hypothesis was that verbs together with nouns should be of paramount importance to the system, as they are the main creators of meaning in language. We find that reversing verb meaning disturbs system output in 9.5% of cases, with little influence on decision certainty. We then proceed to explain this phenomenon by observing the behavior of deep net architecture and the characteristics of the SQuAD dataset (Rajpurkar et al., 2016) itself.

As a basis of our research we use the QA system described in Chen et al. (2017). We pick this model for its good performance and state-of-the-art approach.

## 2 Negating Question Meaning

The first step we take is to measure the impact of verb meaning in question on system output. **First**, we swap verbs in questions for their antonyms using WordNet (Miller, 1995). For auxiliary verbs, we insert their negations (e.g. *is* - *isn't*). If an antonym is not found in WordNet, we substitute a random verb without assuring that its meaning matches the context. Examples of modified questions are presented in Table 1. **Next**, we test how many system outputs for original questions match the outputs for questions with reversed meaning of verbs. As matching we classify identical answers and also some cases with minimal differences in meaning (where we are sure that the system is attending to the same answer), such as *18th overall* vs. *18th*, or *School of Architecture* vs. *Notre Dame School of Architecture*. The test was conducted on SQuAD development set.

In total, we obtained an accurate match (no system decision change) in 90.5% of all tested cases. Mean decision certainty expressed in softmax stayed similar at 0.60 for modified questions and 0.61 for original questions.

## 3 Experiments

Attempting to understand the behavior of the system we take inspiration from works focusing on visualizing deep net internals (Li et al., 2015; Karpathy et al., 2015). We apply measures specific to the mechanisms present in our tested system: question self-attention and hidden layers of the RNN. We run experiments on SQuAD development set.

**Question self-attention** As described in Chen et al. (2017), question self-attention learns to encode importance of each question word. We inspect attention scores $b_j$ for each token during prediction and measure averaged absolute scores for words. As shown in Table 2, indeed question attention learned to devalue verbs. Statistical importance of differences between distributions (in particular, of verbs vs. nouns) was confirmed with Kolmogorov-Smirnov test, which showed *p*-values smaller than 0.001.

---
[*]Both authors contributed equally.

| Original question | Question with verb antonym |
|---|---|
| **Q:** How many teams **participate** in the Notre Dame Bookstore Basketball tournament? | **Q:** How many teams **drop out** in the Notre Dame Bookstore Basketball tournament? |
| **Q:** Which art museum **does** Notre Dame administer? | **Q:** Which art museum **doesn't** Notre Dame administer? |

Table 1: Examples of sentences obtained with inserting verb antonyms.

| PoS | Attention |
|---|---|
| Total Verbs | 2.32 |
| Total Nouns | 5.43 |
| Other PoS | 2.39 |
| AUX Verbs | 0.63 |
| Non-AUX Verbs | 4.16 |
| Non-NE Nouns | 5.21 |
| NE Nouns | 5.83 |

Table 2: Average absolute attention scores for parts of speech. We show scores for all verbs, all nouns, all PoS other than nouns and verbs, auxiliary verbs (AUX Verbs), all verbs other than auxiliary (Non-AUX Verbs), all nouns which are not named entities (Non-NE Nouns) and nouns which are named entities (NE Nouns).

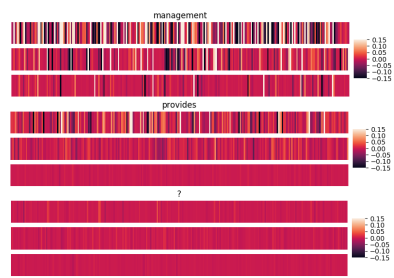

Figure 1: Visualization of values in LSTM hidden layers for a noun (top), verb (middle) and question mark (bottom). Each heatmap shows the 256 values returned by each layer, in 3 layers for each word.

**Hidden LSTM Layers.** Next, we analyze 3-layer LSTM RNN, whose outputs are used to compute question attention. We gather the outputs of all layers and visualize them using heatmaps, as in Figure 1. We observe that variances in numbers appearing in lower layers are distinctively smaller than in the third layer. Furthermore, nouns (in particular named entities) exhibit greater variances than other parts of speech, which aligns with observations for attention scores. Indeed, correlation between entropy scores counted for last hidden layer vectors and attention scores equals -0.91 Pearson's *r*, and and appropriate variance-attention correlation equals 0.85 Pearson's *r* and 0.96 Spearman's correlation, as displayed in Figure 2. It suggests that importance of parts of speech is encoded already by the LSTM network.

**Diagnosis of Dataset.** We observe that in fact the system correctly aligned to the characteristics of the contexts appearing in SQuAD. Most often a specific noun (commonly a named entity, or a combination thereof) appears in a single sentence in single context, so contrasting verbs is not needed to extract the answer. To combat this problem, enhancement of the dataset would be ne-eded to include more sentences with repeating nouns (subjects and objects) and varying verbs describing their actions and relations.

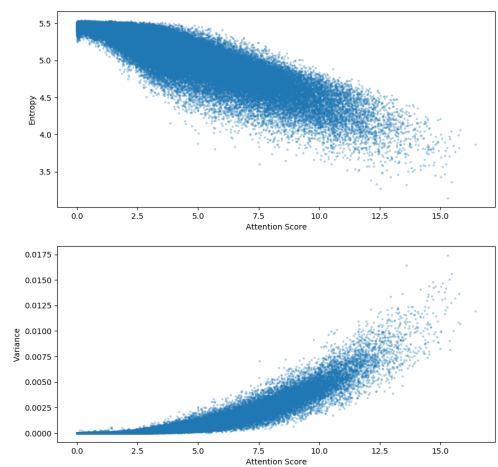

Figure 2: Scatterplot of entropy (top) and variance scores (bottom) in hidden layers (y-axis) and absolute attention scores (x-axis).

**Summary.** We observe low importance of verbs in QA system training on SQuAD dataset and identify shortcomings in the underlying data. Our findings have confirmation in values yielded by network itself. We show that values in hidden layers and attention scores are correlated with importance of words in the question. This work confirms the usefulness of visualization and explanation of deep learning NLP models.